\documentclass[12pt]{article}
\usepackage{amsfonts}
\usepackage{amstext}
\usepackage{comment}
\usepackage{mathtools}
\usepackage{amsmath}
\usepackage{amsthm}
\usepackage{enumitem}
\usepackage[round]{natbib}
\usepackage{hyperref}
\hypersetup{
	colorlinks=true,
	linkcolor=blue,
	citecolor=blue,
	filecolor=magenta,      
	urlcolor=cyan,
}
\usepackage{setspace}
\usepackage{fancyhdr}
\usepackage{float}
\usepackage[dvips]{color}
\usepackage[pdftex]{graphicx}
\numberwithin{equation}{section}
\numberwithin{table}{section}
\usepackage[dvipsnames]{xcolor}
\usepackage{multirow}
\usepackage{subcaption}
\usepackage{subfloat}
\usepackage{commath}
\usepackage{bbm}
\usepackage{longtable}
\usepackage{algorithm}
\usepackage[noend]{algpseudocode}
\usepackage{fullpage}
\usepackage{times}
\usepackage{newtxtext,newtxmath}
\usepackage{authblk}
\usepackage{marginnote}
\usepackage{pbox}

\graphicspath{{figures/}, {other_figures/}}

\newcommand{\argmin}{arg\,min}

\title{
	\bf
	Tree-Based Machine Learning Methods For Vehicle Insurance Claims Size Prediction
}

\author{Edossa Merga Terefe $^{1, 2}$}

\date{$^1$ Research Center for Statistics, University of Geneva, Switzerland \\ %
	$^2$ Statistics Department, Hawassa University, Ethiopia \\ [1ex]%
	\textit{edossa.terefe@unige.ch} \\ [2ex]%
	\today
}

\begin{document}
	
	\maketitle
	
	\section*{Abstract}
	Vehicle insurance claims size prediction needs methods to efficiently handle these claims. Machine learning (ML) is one of the methods that solve this problem. Tree-based ensemble learning algorithms are highly effective and widely used ML methods. This study considers how vehicle insurance providers incorporate ML methods in their companies and explores how the models can be applied to insurance big data. We utilize various tree-based ML methods, such as bagging, random forest and gradient boosting, to determine the relative importance of predictors in predicting claims size and to explore the relationships between claims size and predictors. Furthermore, we evaluate and compare these models' performances. The results show that tree-based ensemble methods are better than the classical least square method.
	
	{\it Keywords:}  claims size prediction; machine learning; tree-based ensemble methods; vehicle insurance.
	
	\section{Introduction} \label{intro}
	A key challenge for the insurance industry is to charge each customer an appropriate price for the risk they represent. Risk varies widely from customer to customer, and a deep understanding of different risk factors helps predict the likelihood and cost of insurance claims. Thus, insurance companies must have an insurance premium that is appropriate for each customer. There are two groups in the insurance industry: life insurance and non-life insurance. This study considers nonlife insurance, particularly vehicle insurance. Insurance claims occur when the policyholder creates a formal request to an insurer for coverage or compensation for an unfortunate event of an accident. Policyholders can mitigate the costs involved with coverage for the property (damage or theft to a car) and liability (legal responsibility to others for the medical or property costs).
	
	Insurance companies must predict how many claims are going to occur and the severity of these claims to enable insurers to set a fair price for their insurance products accordingly. In other words, claim prediction in the vehicle insurance sector is the cornerstone of premium estimates. Furthermore, it is crucial in the insurance sector to plan the correct insurance policy for each prospective policyholder.
	
	Several studies have been done to personalize the premium estimate, such as \cite{guillen2019} and \cite{roel2017}, they demonstrated the possible benefits of analyzing information from telematics when determining premiums for vehicle insurance. The predictive capacity of covariates obtained from telematics vehicle driving data was investigated by \cite{goa2018} and \cite{goa2019} using the speed–acceleration heat maps suggested by \cite{wuthrich2017}.
	
	Prediction accuracy enables the insurance industry to better adjust
	its premiums, and makes vehicle insurance coverage more affordable for more drivers. Currently, many insurance companies are transitioning to ML techniques to predict claims size.  However, selecting a suitable ML predictive model is far from trivial. In this study, we investigate flexible ML techniques to make accurate predictions for claims size by analyzing a large vehicle dataset given by Ethiopian Insurance Company, one of the main car insurance company based in Ethiopia, and we apply the tree-based ML methods to the dataset, such as bagging, random forest, and gradient boosting. We also evaluate and compare the performance of these models.
	
	The rest of this paper is organized as follows.  In Section~\ref{data}, the dataset is described and some descriptive statistics are provided. In Section~\ref{background}, we present review of three tree-based ensemble methods is presented. In Section~\ref{results}, we report the results from application of considered methods. In Section~\ref{conclusion} we provide a discussion and conclusion of the study.

	\section{Dataset and Exploratory Analysis} \label{data}
	\subsection{The data}
	The data used for this analysis were provided from a large database of the Ethiopian Insurance Corporation, one of the biggest insurance companies in Ethiopia. It consists of policy and claim information of vehicle insurance at the individual level. The dataset originally contains $n= 288,763$ unique individual contracts, represented by the observations $(X_1, Y_1), \dots ,(X_n, Y_n)$ where $X = (X_1,\dots,  X_p)\in \mathbb R^p$ denotes a vector of $p=10$ predictors, and $Y \in \mathbb{R}$ denotes the response variable representing the claim size. The data were correspond to the period between July 2011 to June 2018. The different predictors used in the analysis are summarized in Table~\ref{table 1}.
	
	\begin{table}
		\begin{center}
			\begin{tabular}{||l|l|l|l|l||} \hline \hline
				S.N & Name            & Type        & Domain / Levels                     & Description / representation                   \\ \hline \hline
				1   & Sex             & categorical & 0, 1, 2                             & \pbox{20cm}{0 = legal entity, 1 = male,        \\ 2 = female} \\  \hline
				2   & Season          & categorical & \pbox{20cm}{autumn, winter, spring,                                                  \\ summer} & Beginning of contract. \\ \hline
				3   & Insurance type  & categorical & 1201, 1202, 1204                    & \pbox{20cm}{1201 = private, 1202 = commercial, \\ 1204 = motor trade road risk} \\ \hline
				4   & Type vehicle    & categorical & pick-up, truck, bus, ...            & \pbox{20cm}{Type of vehicle grouped into six   \\ categories.} \\ \hline
				5   & Usage           & categorical & \pbox{20cm}{fare paying passengers,                                                  \\ taxi, general cartage, ...} & \pbox{20cm}{A usual usage of the vehicle \\ grouped into six categories.} \\ \hline
				6   & Make            & categorical & Toyota, Isuzu, Nissan,...           & Manufacturer company.                          \\ \hline
				7   & Coverage        & categorical & comprehensive, liability            & Scope of the insurance.                        \\ \hline
				8   & Production year & Integer     & 1960 - 2018                         & Vehicle's production year.                     \\ \hline
				9   & Insured value   & continuous  & $\mathbb{R+}$                       & Vehicle's price in USD.                        \\ \hline
				10  & Premium         & continuous  & $\mathbb{R+}$                       & Premium amount in USD.                         \\ \hline \hline
			\end{tabular}
			\caption{Description of predictors in Ethiopian vehicle insurance dataset.}
			\label{table 1}
		\end{center}
	\end{table}
	
	The terms, liability and comprehensive \texttt{coverage} in Table \ref{table 1} are defined as:
	\begin{itemize}
		\item Comprehensive \texttt{coverage}: The company covers all the losses which happen to the car whenever the conditions of agreement are satisfied.
		\item Liability or third party \texttt{coverage}: The car can cause a damage to someone or someone's property. If the policy holder already has an agreement for a liability coverage, the insurance company covers the costs in this case. Liability amount, which a policy holder has to pay as a part of premium is nationally fixed almost every year for each car type across all the insurance companies operating in the country. The liability cases are usually taken to courts or settled by negotiation. However, the affected party should not be either a family member or a relative of the policy holder.
	\end{itemize}
	
	Computation of premium is determined as a function of:
	
	\begin{itemize}
		\item Insured value.
		
		\item Production year: For the first three years, age of the car is not considered, but after three years, age loader computation technique, which takes into account age of the vehicle is applied.
		
		\item No claim discount (NCD): Upon the renewal of the contract after a year, a policy holder gets 20\% discount from the previous year's premium and adjusted for inflation if he/she has not applied for a claim in that year. He/she can get up to 60\% of discount in the consecutive years but pooled down by the age loader of the vehicle.
		
		\item Contingency, Plant and Machinery (CPM): Applicable for those cars which are operated on different circumstances. For instance, loaders type vehicles premium can be computed depending on the assumption of being at the engineering (construction) sites. But in case the vehicle cause an accident while being driven on the road, the computation needs an additional consideration and computation mechanism differs.
	\end{itemize}

	Some predictors such as carrying capacity and seat number are removed from the dataset prior to data analysis and modeling since they are not correctly coded.

	\subsection{Claims size Variable}
	
	In our analysis, the claim size is a continuous response variable $Y \in \mathbb{R}$. It is originally the amount in the Ethiopian currency Birr and it is converted to USD during data analysis. The distribution of the response variable $Y$ is strongly zero-inflated since for about $92.5\%$ of the contracts there is no claim paid. Thus, instead of visualizing and modeling this distribution directly, we first select the non zero observations. Given that the claim has to be paid for policy holder $i$, it is determined as

	\begin{equation}\label{eq:y}
		Y_i = \text{Claims Size} = \frac{\text{Insured value}_i}{\text{Market value}_i}\times \text{Loss}_i,
	\end{equation}

	where
	\begin{itemize}
		\item Market value is the market price of the car when it was bought. The market price for each car type is collected by the insurer almost every year. The data on the market value are taken either from importers or informally from other institutions. It can be either more or less than the insured value. Knowing market value of the car helps to adjust the claim amount not to be too high in case insured value is not reliable. In most cases, insured value and market value are the same.
		
		\item Loss: When an accident happens, the damaged vehicle is inspected by the engineer experts who work in the insurance company. These experts known as server decision look at each part of the vehicle, identify the affected parts and propose either to replace or fix the vehicle.  Once the affected parts have been identified, its price is determined by the server decision and a bid for repairing the damaged car is published. Including the server decision members, anyone who has a license to do so can usually participate in the bid competition. The loss is determined as follows.
	\end{itemize}

	In some cases, the amount of claim paid can be higher than either the insured value or market price. It is mandatory to have at least a liability insurance coverage for all vehicles as a country's regulation, even if comprehensive insurance coverage in for a safety of the vehicle's owner. Additionally, a policy holder can also have BSG and PLL insurance coverage, but if only the comprehensive coverage is already secured first. The terms BSG and PLL are defined as:
	
	\begin{itemize}
		\item Bandit, Shifta and Gorilla (BSG):  A contract agreement in case the car is robbed or stolen. To the maximum of the insurer's knowledge, it is a vehicle insurance component applicable in Ethiopia only.
		
		\item Passengers legal liability (PLL): This is applicable for fare paying passengers, in case someone is affected by an accident being in the car. Similar to that of liability coverage, its amount is fixed to be paid as part of premium and a maximum of 40,000 birr would be paid by the insurer to a passenger in case an accident happened.
	\end{itemize}
	
	Even though both BSG and PLL insurance coverage depend on the interest of policy holder and they are optional, applicable if and only if the comprehensive insurance agreement is to take place or has already taken place.\vspace{0.25cm}

	The insured value in \eqref{eq:y} does not include the values of liability, BSG and PLL, even though they are included in the contract. It contains the value of comprehensive coverage only. Thus, claims size can be higher than the insured value if:
	
	\begin{itemize}
		\item (total) loss + (liability insurance) + (PLL) $>$ insured value,
	\end{itemize}
	where total loss is an overall loss of the car due to a severe accident and impossible to repair. In that case the insurer company pays exactly the insured value as a claim.
	
	\subsection{Exploratory Data Analysis} \label{eda}
	
	To make assumptions about the data and find a model that fits it best, it is important to carry out an exploratory data analysis (EDA), since it has a significant role to let the data speak for themselves prior to or as part of a formal analysis. It allows the researcher to influence and criticize an intended analysis. Additionally, EDA techniques may reveal additional information that may not be directly related to the research question. For example, EDA could suggest fruitful new lines of research \citep{maind2010}.
	
	The purpose of statistical graphics is to provide visual representations of quantitative and qualitative information. As a methodological tool, statistical graphics comprise a set of strategies and techniques that provide the researchers with important insights about the data under examination and help guide for the subsequent steps of the research process. The objectives of graphical methods are to explore and summarize the contents of large and complicated data sets, address questions about the variables in an analysis (for example, the distributional shapes, ranges, typical values and unusual observations), reveal structure and pattern in the data, check assumptions in statistical models, and facilitate greater interaction between the researcher and the data. Various graphical methods were examined to visualize data in raw and amalgamated formats.
	
	The most widely recognized graphical tool to display and examine the frequency distribution and a density of a single continuous variable is the histogram.
	
	\begin{figure}[H]
		\begin{center}
			{\includegraphics[scale=1]{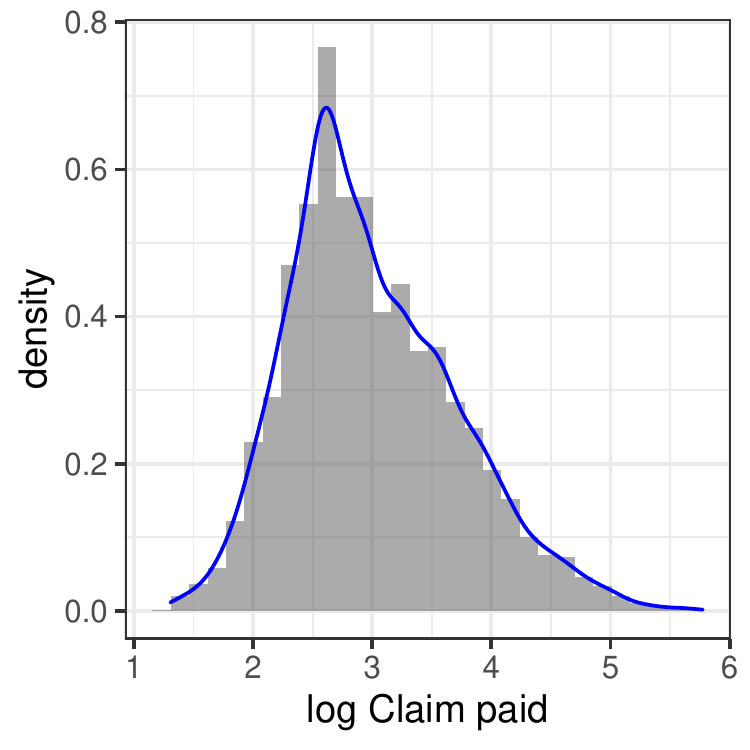}}
			\caption{Frequency histogram and superimposed density plot representations of natural logarithm of claim paid distribution. The distribution of the response variable $Y$ is strongly zero-inflated since for only about $7.5\%$ of the contracts there is non-zero claim payment. Thus in our analysis, instead of visualizing and modeling this distribution directly, we only consider a data of policy-holders who have ever received a positive claim.}
			\label{fig: hist}
		\end{center}
	\end{figure}
	
	Another common tool to visualize the observed distribution of data is by plotting a smoothed histogram commonly referred as empirical density, the blue curve superimposed on the histogram with blue line in Figure \ref{fig: hist}.  The empirical densities overcome some of the disadvantages caused by the arbitrary discrete bins used in the basic histograms.
	
	\subsection{Exploring relationships between covariates and Claim paid}
	
	Relationships between the predictors and the response variable can be depicted by graphical methods. Side-by-side boxplots are one way of graphical displaying the relationship between qualitative and quantitative variables. It is an excellent tool for conveying location and variation information in data sets, particularly for detecting and illustrating location and variation changes between different groups of data.
	
	\begin{figure}[!ht]
		\begin{center}
			{\includegraphics[width=1\textwidth]{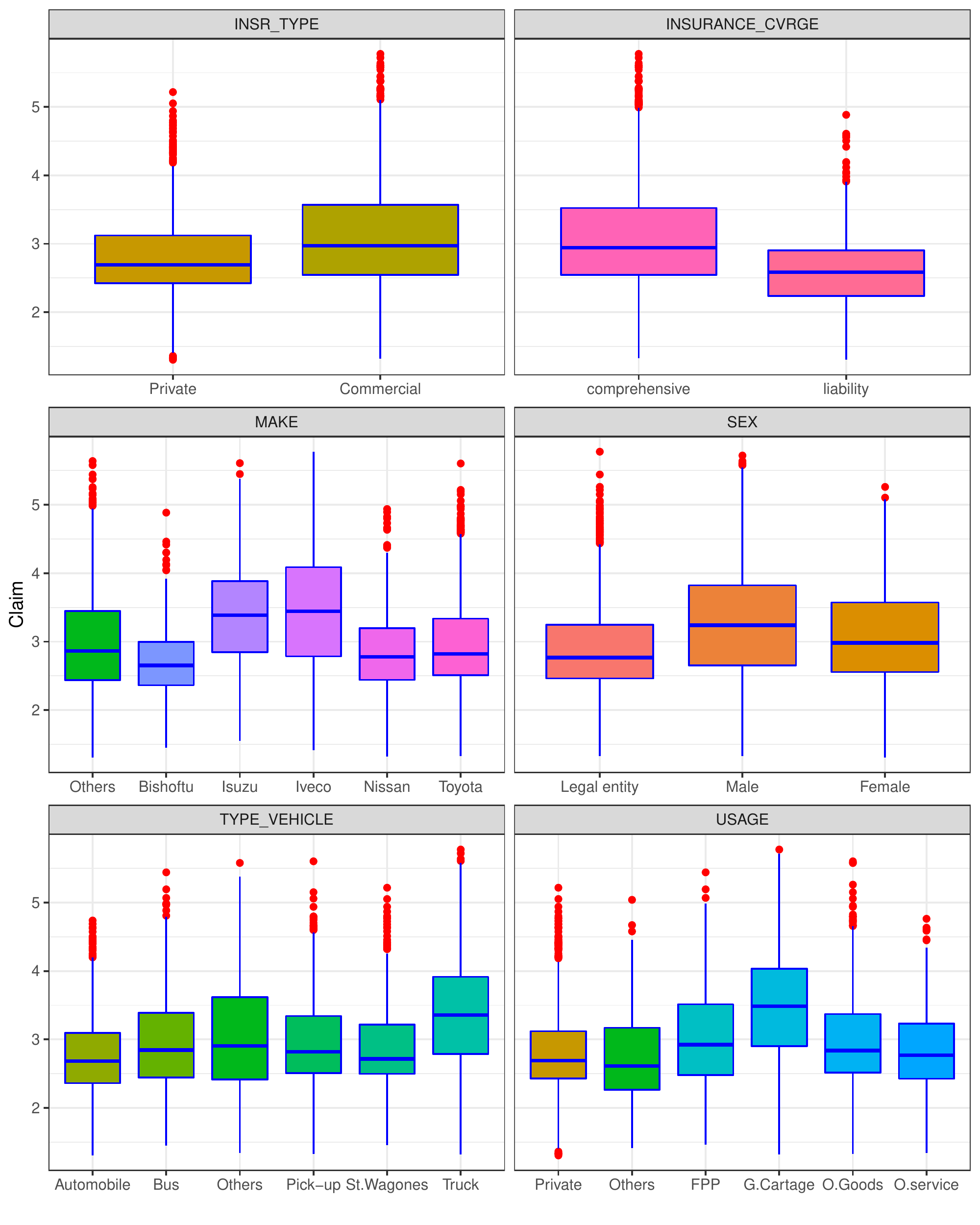}}
			\caption{Boxplots of natural logarithm of claim paid against qualitative predictors.}
			\label{fig:boxplots}
		\end{center}
	\end{figure}
	
	The boxplots of claim paid against the different qualitative predictors are shown in Figure \ref{fig:boxplots}. Several predictors seem to have significant heterogeneity across theirs labels. For instance, in the boxplots of log of claim paid against sex, it seems existence of differences in terms of log of claim paid across the three groups of sex. Accordingly, male policy holders appeared to have a higher claim payment than either female counterparts or legal entities. Similarly, differences in claims size are observed in vehicle usage, identifying the vehicles that used for a general cartage to have the highest claim payment followed by a fare paying passengers vehicles. Moreover, vehicles manufactured from Isuzu and Iveco companies cost the insurer more than vehicles that are from any other companies, which is consistent with the insurer's prior identification of risky vehicles. It can also be seen that there are differences across the groups in other covariates such as vehicle type, insurance type and insurance coverage.
	
	Boxplots are also a robust measure of the spread of a distribution and more informative than merely computing the variance by group as they can be helpful in identifying the homogeneity of variance between groups of a predictor. Looking at the boxplot of sex covariate again, it can be seen that the claim payment made for male policy holders appears to be more variable than either of the other two categories. And also in vehicle usage covariate, claim payment made for vehicles that are used for a general cartage and fare paying passengers purposes have more variability than any other groups. Similarly, heterogeneity of variance between a groups of insurance coverage, insurance type, manufacturer company and vehicle type covariates was observed.
	
	Analogous to boxplots, Scatterplots are an obvious way to visualize a relationship between two quantitative variables that are numerically comparable. They are useful as a preliminary step before performing a regression analysis.
	
	\begin{figure}[!ht]
		\begin{center}
			{\includegraphics[width=1\textwidth]{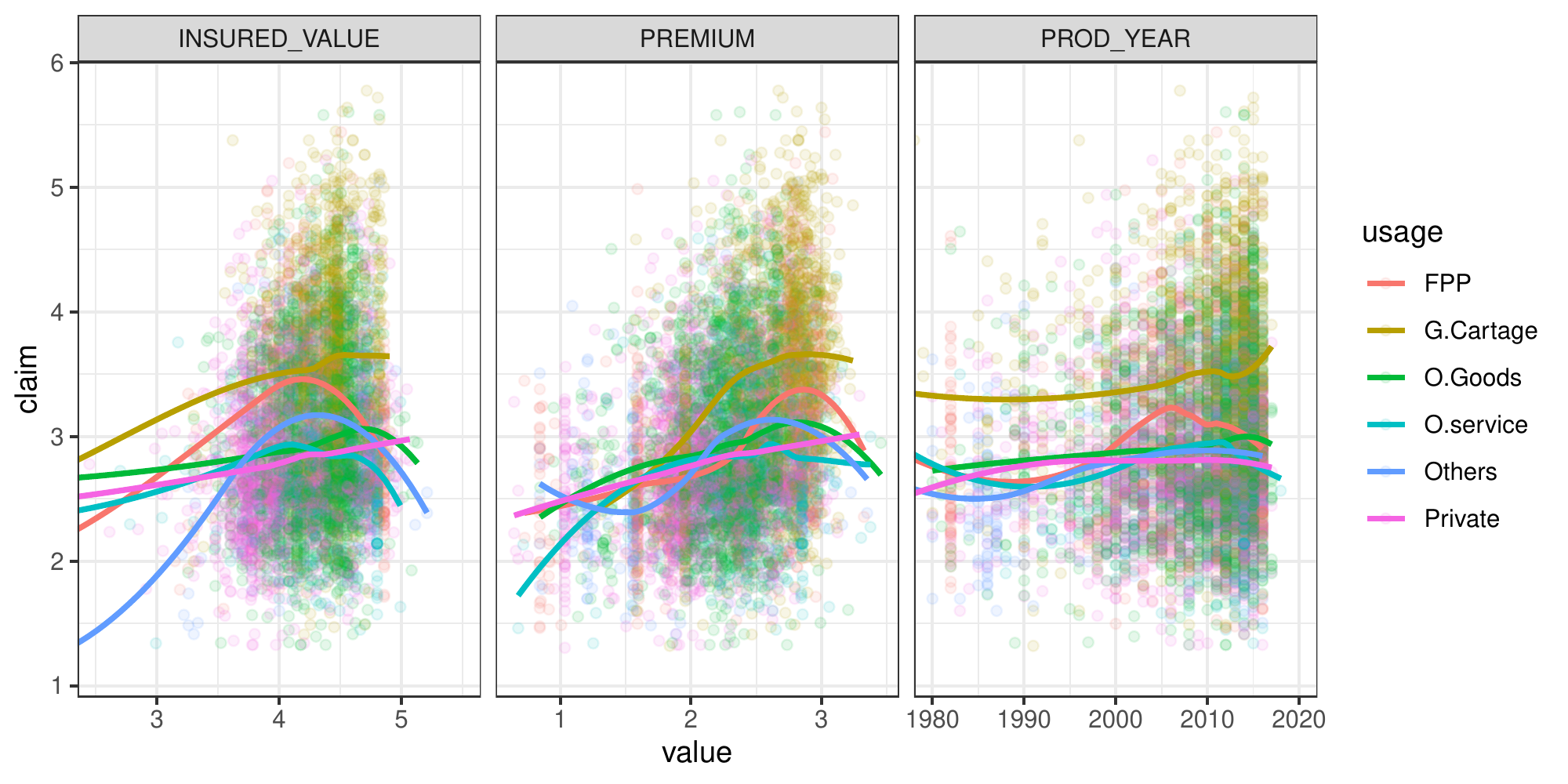}}
			\caption{Scatterplots matrix: a bivariate profiling of relationships between claim and quantitative predictors.}
			\label{fig:scatterplots}
		\end{center}
	\end{figure}
	
	Figure \ref{fig:scatterplots} shows that scatterplots, a bivariate relationships between claim paid and quantitative predictors. It is difficult to detect clear trends in any of the plots. However, by stratifying the points according to the different groups of usage predictor, we see some differences in claims size across the groups.
	
	In addition to the scatterplot matrix seen in Figure \ref{fig:scatterplots}, we computed correlation coefficients between claims size and insured value, premium and production year as $0.22$, $0.33$ and $0.11$, respectively. Even though none of the coefficients between claims size and the covariates are considered to be strong, but there are some notable associations. For instance, claims size appear to have a moderate positive correlation between insured value, premium and production year , meaning that as vehicles' insured value, premium and production year increase, their claim payment also tends to increase. Correlation coefficient based relationships usually be teased out more clearly when building the (final) model.
	
	The \textit{loess curves} drawn on top of the scatterplots indicates a possibly nonlinear relationship between the two variables. The curves for claims size against insured value and premium are an upside-down U-shape, peaking around the middle of insured value and premium for the most groups of usage predictor. This means that the vehicles with moderate insured value and/or moderate premium have larger claims sizes than those with lower and higher insured value and /or premium. Because this trend is non-linear, this finding could not have been inferred from the correlations alone. On the other hand, when we consider the private group of usage predictor, the relationships between claim size against insured value and premium seem to be linear with a positive slope. 
	
	\section{Review of Machine Learning Methods } \label{background}
	Machine learning is now well established in many areas. In contrast to the statistical modeling approach, ML algorithms do not assume any specific model structure for the data. ML methods capture the underlying structure of data and therefore, they are more efficient in handling large data with arbitrary degree of complexity. One major task of machine learning is to construct good models from data sets.
	
	Among ML algorithms, ensemble methods are one of the usual choice to analyze large and complex data. Originally developed to reduce the variance-thereby improving the accuracy of an automated decision-making system, ensemble methods have since been successfully used to address a variety of machine learning problems \citep{zhang2012}, such as predictor selection, class imbalanced data, confidence estimation, error correction, among others.
	
	The main idea behind the ensemble methodology is to weigh several individual pattern learners, and combine them in order to obtain a learner that outperforms most of them. In fact, combining the learners outputs does not necessarily lead to a performance that is guaranteed to be better than the best learner in the ensemble. Rather, it reduces likelihood of choosing a learner with a poor performance. Ensemble methodology imitates to seek several opinions before making any crucial decision. The individual opinions are weighted, and combined to reach the final decision \citep{poli2006}.
	
	A general principle of ensemble methods is to construct a linear combination of some model fitting method, instead of using a single fit of the method to improve the predictive performance of a given statistical learning or model fitting technique. More precisely, consider an estimation of a real-valued function $f : \mathbb{R}^p \rightarrow \mathbb{R}$ based on data $\left\{(x_i,y_i); i = 1, \dots, n\right\}$ where $x$ is a $p$-dimensional predictor variable and $y$ a univariate response. We may then generalize to functions $f(x)$ and other data types.
	
	Given some input data, we learn several functions $\hat{f}_1, \hat{f}_2, \hat{f}_3, \dots, \hat{f}_B$, called learners, by changing the input data based on different reweighting. We can then construct an ensemble-based function estimate $\hat{f}_{ens}(x)$ by taking linear combinations of the individual learners as an additive expansion of the learners \citep{elish2009} $\hat{f}_{i}(x)$:
	
	\begin{equation}\label{ens}
		\hat{f}_{ens}(x) = \sum_{i = 1}^{B} w_i \hat{f}_i(x),
	\end{equation}
	
	where the $\hat{f}_i(x)$ are estimates obtained from the $i^{th}$ reweighted dataset and $w_i$ are the linear combination coefficients. For instance, $w_i = 1/B$, an averaging weights for bagging (see Section~\ref{bagging_sec}) and for boosting (see Section~\ref{boosting_sec}).
	
	In this study, three ensemble learning algorithms i.e., bagging, random forest and boosting are considered. For models performance comparison purpose, two non-ensemble learning technique i.e., ordinary linear regression and decision tree are also applied to Ethiopian vehicle insurance data set to predict claims size.

	\subsection{Bagging} \label{bagging_sec}
	
	Bagging \citep{brepred1996}, which stands for \textbf{b}ootstrap \textbf{agg}regat\textbf{ing}, is an ensemble method for improving unstable estimation or classification schemes. As the name implies, the two key ingredients of Bagging are bootstrap and aggregation.
	
	Bagging adopts the bootstrap distribution for generating different learners. In other words, it applies bootstrap sampling \citep{efron1993} to obtain the data subsets for training the learners. In detail, given an original data set, we generate a data set containing $n$ number training observations by sampling with replacement. Some original observations appear more than once, while some original observations are not present in the sample. By applying the process $B$ times, $B$ samples of $n$ training observations are obtained. Then, from each sample a learner can be trained by applying the learning algorithm.
	
	Bagging also adopts the most popular strategies for aggregating the outputs of the learners, that is, voting for classification and averaging for regression. To predict a test instance, taking regression for example, Bagging feeds the instance to its learners and collects all of their outputs, and then takes the average of the outputs as the prediction, where ties are broken arbitrarily.
	
	In particular, the bagging method for regression is applied as follows. A learner $\hat{f}_i(x)$ is fitted on each of the $B$ bootstrapped sample, where $\hat{f}_i(x)$ denotes the predicted response values from $i = 1,2, \dots, B$ learners. Then the $B$ learners constructed are combined using the aggregation, so that the average prediction, $f_{av}(x)$ is estimated as the average of predicted outputs from $\hat{f}_i(x)$ as:
	
	\begin{equation}\label{bagging_final}
		\hat{f}_{av}(x) = \frac{1}{B} \sum_{i = 1}^{B} \hat{f}_i(x)
	\end{equation}
	
	in order to obtain a single low-variance learner.

	\subsection{Random Forest} \label{rf_sec}
	
	Random Forest (RF) is a representative of the state-of the-art ensemble methods algorithm developed by \cite{brei2008}, and a very powerful technique which is used frequently in the data science field across industries \citep{dang2017}. It is an extension of Bagging, where the major difference with Bagging is the incorporation of randomized predictor selection. During the construction of a component decision tree, at each split, RF first randomly selects a subset of predictors, and then carries out the conventional split selection procedure within the selected predictor subset.
	
	RF is usually applied to reduce the variance of individual trees by growing numerous trees. Each subsequent split for all trees grown is not done on the entire data set, but only on the portion of the prior split that it falls under. For each tree grown, about one third of training samples are not selected in bootstrap and it is called out of bootstrap ("out of bag" or "OOB") samples, as in case of Bagging in Section \ref{bagging_sec}. Using OOB samples as input to the corresponding tree, predictions are made as if they were novel test samples. A particular observation can fall in the terminal nodes of many trees in the forest, each of which, potentially, can give a different prediction. Again, the OOB sample data used to fit a particular tree is used to make each tree's prediction. Through book-keeping principle, an average for continuous response is computed for all OOB samples from all trees for the prediction of RF model. For discrete outcomes, the prediction is the majority votes from all trees that have been grown without the respective observation or the average of the predicted probabilities \citep{jone2015}.

	\subsection{Gradient Boosting} \label{boosting_sec}
	
	Gradient Boosting algorithms have been proposed in the machine learning literature by \cite{scha1990} and [ \cite{freu1995}; \cite{freu1996}]. The term \textbf{boosting} refers to a family of algorithms that are able to convert weak learners to strong learners for improving the predictive performance of a regression or classification procedure. Intuitively, a weak learner is just slightly better than random guess, while a strong learner is very close to perfect performance. The boosting method attempts to boost the accuracy of any given learning algorithm by fitting a series of models, each having a low error rate, and then combining them into an ensemble that may achieve better performance \citep{scha1999}. This strategy can be understood in terms of other well-known statistical approaches, such as additive models and a maximum likelihood \citep{fried2000}.
	
	Like bagging, boosting is a general approach that can be applied to many ensemble statistical learner for regression or classification. Unlike bagging which is a parallel ensemble method, boosting methods are sequential ensemble algorithms where the weights $w_i$ in \eqref{ens} are depending on the previous fitted functions $\hat{f}_1, \dots, \hat{f}_{i-1}$. Boosting does not involve bootstrap sampling; instead each tree is fitted on a modified version of the original data set.
	
	There are several versions of the boosting algorithms for classification problems [\cite{druc1997}; \cite{fried2000}; \cite{scha2013}], but the most widely used is the one by \cite{freu1996}, which is known as AdaBoost, and it has been empirically demonstrated to be very accurate in terms of classification.
	
	There are also several studies [\cite{fried2002}; \cite{frie2001}; \cite{druc1997}] conducted related to boosting for regression problems. In this paper, we rely on a recently proposed gradient boosting algorithm by \cite{chen2016_xgboost}, which uses regression trees as the basis functions, and it optimizes a regularized learning objective function
	
	\begin{equation}
		\begin{aligned}\label{eq:object_boost}
			L(\phi)      & = \sum_{i} l(\hat{y}_i, y_i) + \sum_{b}\Omega(f_b)              \\
			\text{where} & \hspace{0.25cm} \Omega = \gamma T + \frac{1}{2}\lambda ||w||^2.
		\end{aligned}
	\end{equation}
	
	Here, $l$ is a differentiable convex loss function that measures the difference between the prediction $\hat{y}_i$ and the target $y_i$. The second term $\Omega$ penalizes the complexity of the model (i.e., the regression tree functions). The additional regularization term helps to smooth the final learned weights to avoid over-fitting.
	
	Boosting regression tree involves generating a sequence of trees, each grown on the residuals of the previous tree. Therefore, boosting regression tree model inherits almost all of the advantages of tree-based models, while overcoming their primary disadvantage, that is, inaccuracy \cite{frie2003}.
	
	\section{Application of Machine Learning Methods to Insurance Data} \label{results}
	
	\subsection{Variable Importance} \label{subsec:vip}
	
	Our goal is not only to find the most accurate model of the response, but also to identify which of the predictor variables are most important to make the predictions. For this reason, we perform variable importance. The ensemble methods algorithm estimate the importance of for instance, $x_j$ predictor variable by looking at how much prediction error increases over the baseline error, when the OOB sample for $x_j$ predictor is permuted while all others are left unchanged.  The most commonly used variable importance measure is the permutation importance, introduced by \cite{brei2001}, which suggests that the variable importance of predictor $x_j$ is the difference in prediction accuracy after and before permuting $x_j$ averaged over all trees.  More precisely, the variable importance of predictor $x_j$ is defined as
	
	\begin{equation}\label{eq:vip}
		\text{VI} (x_j) = \frac{1}{B}\sum_{i=1}^{B} \text{VI}\left(x_j\right)_i
	\end{equation}
	
	where $\text{VI}\left(x_j\right)_i = \left(\text{RMSE}_i^{x_j} - \text{RMSE}_i^{0}\right)$, and $\text{RMSE}_i^{x_j}$ representing the $\text{RMSE}$ value for the $i^{th}$ model in the ensemble fitted from the dataset with a random permutation applied to the covariate $x_j$, and $\text{RMSE}_i^0$ is the $\text{RMSE}$ value for this model fitted to the original dataset. Note that $\text{VI}\left(x_j\right)_i = 0$, if variable $x_j$ is not in the $i^{th}$ model. The raw variable importance score for each variable is then computed as the mean importance over all trees. In fact, $\text{VI}(x_j)$ can be computed depending on any other performance measure such as coefficient of determination, $R^2$.
	
	Let $x_1, \dots, x_p$ be the features of interest and let $\text{RMSE}_0$ be the baseline performance metric for the trained model. The permutation-based variable importance scores can be computed as shown in Algorithm \ref{alg:vip}.
	
	\begin{algorithm}
		\caption{Permutation-based variable importance computation.}
		\label{alg:vip}
		\begin{algorithmic}[1]
			\State \textbf{For} $j \in \left\{1, 2, \dots, p\right\}$;
			\begin{itemize}
				\item[(a)] Permute the values of feature $x_i$ in the training data.
				\item[(b)] Recompute the performance metric on the permuted data, $\text{RMSE}$.
				\item[(c)] Record the difference from baseline using \ref{eq:vip}.
			\end{itemize}
			\State Return the variable importance scores $\text{VI}(x_1), \dots, \text{VI}(x_p)$.
		\end{algorithmic}
	\end{algorithm}
	
	\begin{figure}[H]
		\begin{center}
			{\includegraphics[width=1\textwidth]{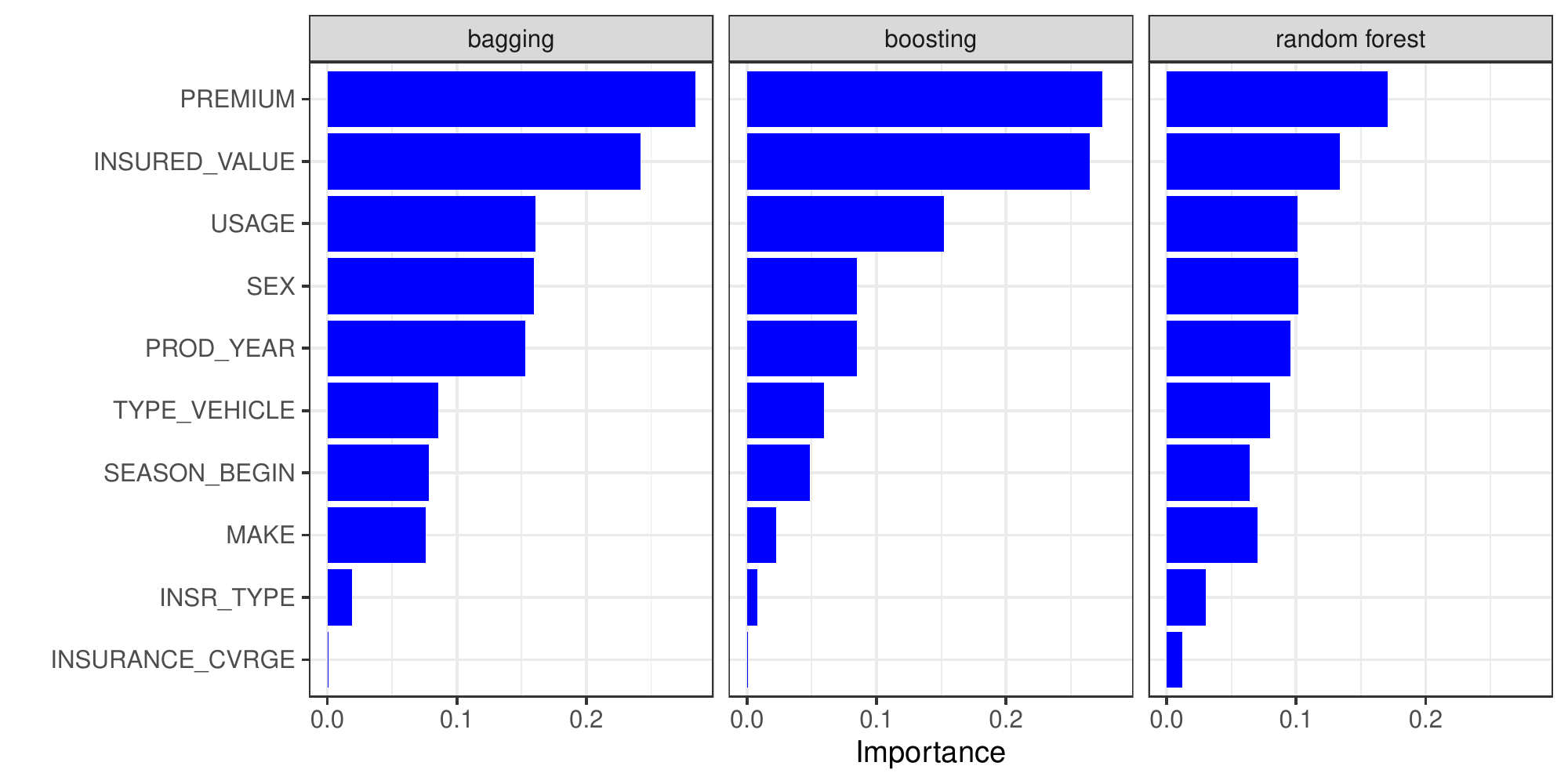}}
			\caption{A graphical representation of average variable importance across all the trees from Bagging, boosting and regression RF. The larger the number, the bigger the effect.}
			\label{fig:vip}
		\end{center}
	\end{figure}
	
	Figure \ref{fig:vip} displays the importance predictor while growing the trees.  Accordingly in all the three models, premium is the most crucial predictor followed by the insured value. The second most influential (slightly equally in bagging and random forest) predictors are usage and sex, followed by production year, in line with the earlier boxplots exploratory analysis.
	
\begin{figure}[H]
		\begin{center}
			{\includegraphics[width=0.455\textwidth]{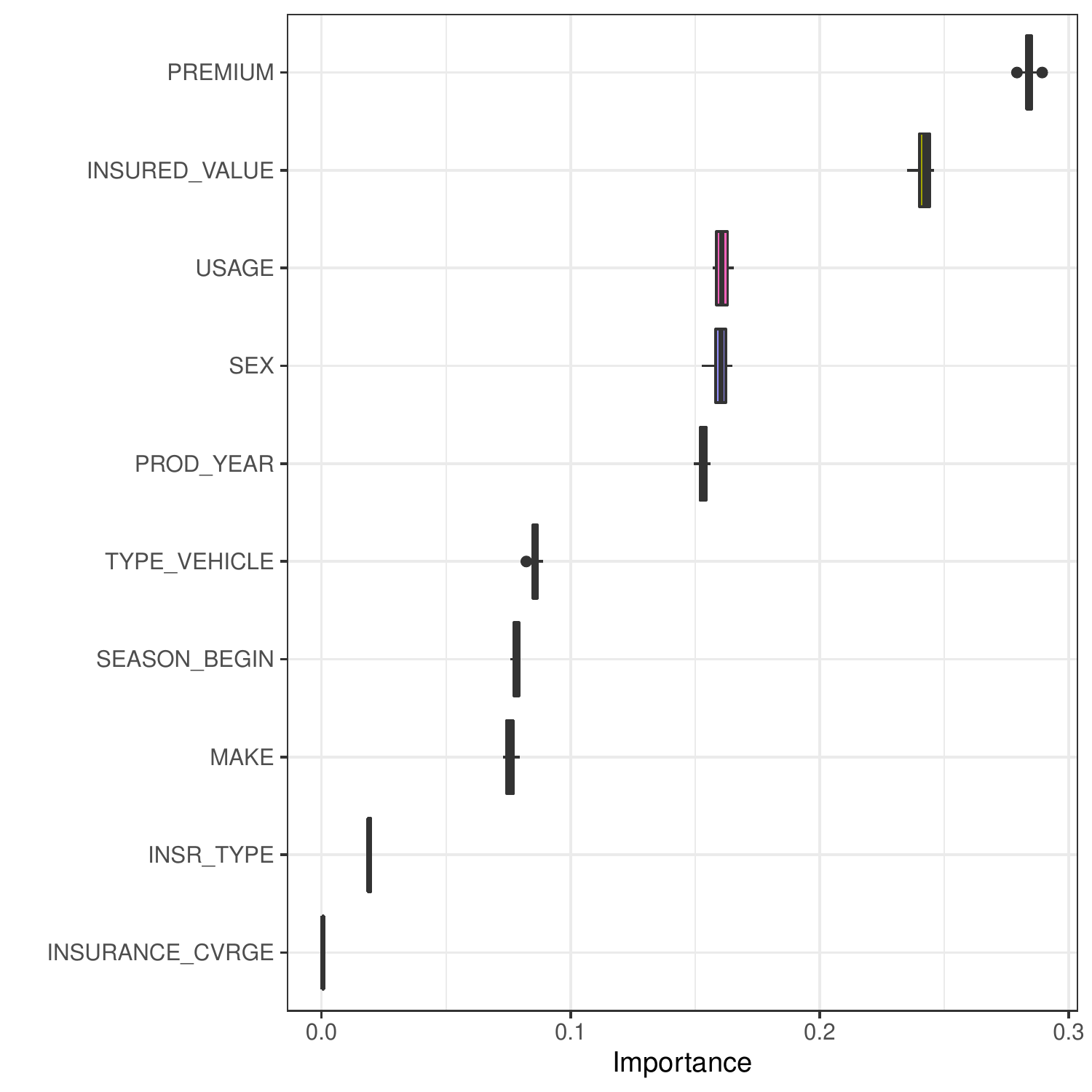}}
			{\includegraphics[width=0.455\textwidth]{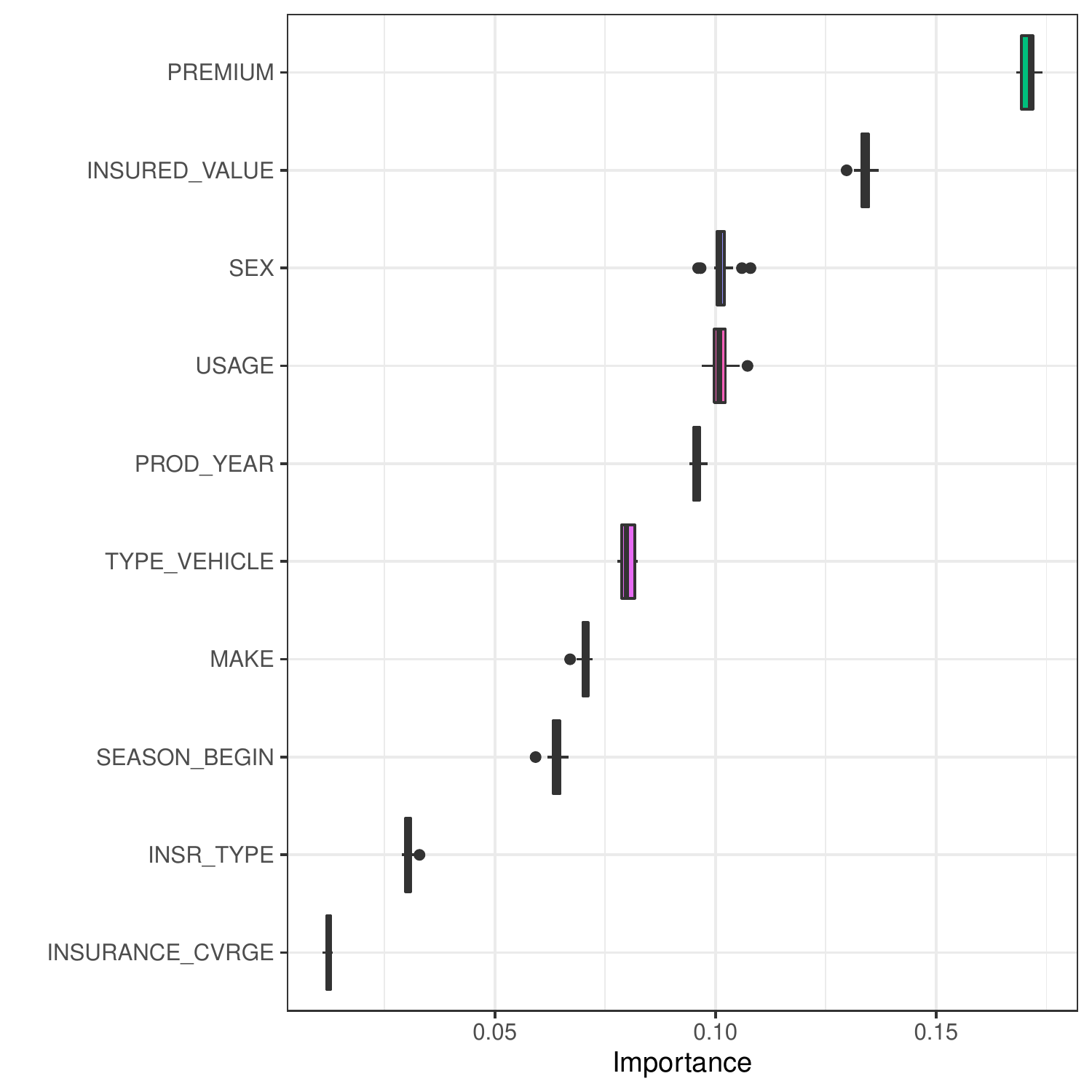}}
			\caption{Boxplots of variable importance from Bagging (left panel) and regression RF (right panel) from $20$ times repeated permutation.}
			\label{fig:vip_box}
		\end{center}
	\end{figure}

	Figure \ref{fig:vip} is obtained by repeating the permutation of each variable $20$ times and the results averaged together. This helps to provide more stable VI scores, and also the opportunity to measure their variability as seen in Figure \ref{fig:vip_box}, since permutation approach introduces randomness into the procedure.
	
	\subsection{Partial Dependence Plots} \label{subsec:pdp}
	
	Though determining predictor importance is a crucial task in any supervised learning problem, ranking variables is only part of the story and once the important predictors are identified it is often necessary to assess the relationship between the predictors and the response variable.
	The task is often accomplished by constructing partial dependence plots (PDP) \citep{frie2001},  which helps to visualize the relationship between a predictor and the response variable while accounting for the average effect of the other predictors in the model.

	Let $\hat{y}$ be prediction function from an arbitrary model using a dataset, $D = \left\{(x_{i,j}, y_i)\right\}$ for $i = 1, \dots, n$ and $j = 1, \dots, p$. The model generates predictions of the form:
	
	\begin{equation}\label{eq:pdp}
		\hat{y}_i = f\left(x_{i,1}, x_{i,2}, \dots, x_{i,p}\right),
	\end{equation}
	for some function $f(...)$.
	
	Let $x_k$ be a single predictor of interest with unique values $\left(x_{1,k}, x_{2,k}, \dots, x_{n,k}\right)$. Then the partial dependence plots are obtained by computing the following average and plotting it over a useful range of $x$ values:
	
	\begin{equation}\label{eq:pdp_single_x}
		\bar{f}_k(x) = \frac{1}{n} \sum_{j = 1}^{n}\hat{f}\left(x_{1,j}, \dots, x_{k-1,j}, x, x_{k+1,j}, \dots, x_{p,j}\right)
	\end{equation}
	The function $\bar{f}_k(x)$ indicates how the value of the variable $x_k$ influences the model predictions $\left\{y_j\right\}$ after we have averaged out the influence of all other variables. The partial dependence of the response on $x_k$ can be constructed as Algorithm \ref{alg:pdp}:
	
	\begin{algorithm}
		\caption{Partial dependence construction of the response on a single predictor $x_k$.}
		\label{alg:pdp}
		\begin{algorithmic}[1]
			\State \textbf{For} $j \in \left\{1, 2, \dots, n\right\}$;
			\begin{itemize}
				\item[(a)] Replace the original training values of $x_k$ with the constant $x_{k,j}$.
				\item[(b)] Compute vector of predicted values, $\left\{\hat{y}_j\right\}$ from the modified version of the training data.
				\item[(c)] Compute the average prediction according to \ref{eq:pdp_single_x} to obtain $\bar{f}_{k}(x_{k,j})$.
			\end{itemize}
			\State Plot the pairs of $\left\{x_{k,j}, \bar{f}_{k}(x_{k,j})\right\}$ for $j = 1, 2, \dots, n$
		\end{algorithmic}
	\end{algorithm}
	
	Since Algorithm \ref{alg:pdp} can be quite computationally intensive as it involves $n$ passes over the training records, a reduced number of points is used by equally spacing the values in the range of the variable of interest.
	
	\begin{figure}[H]
		\begin{center}
			{\includegraphics[width=1\textwidth]{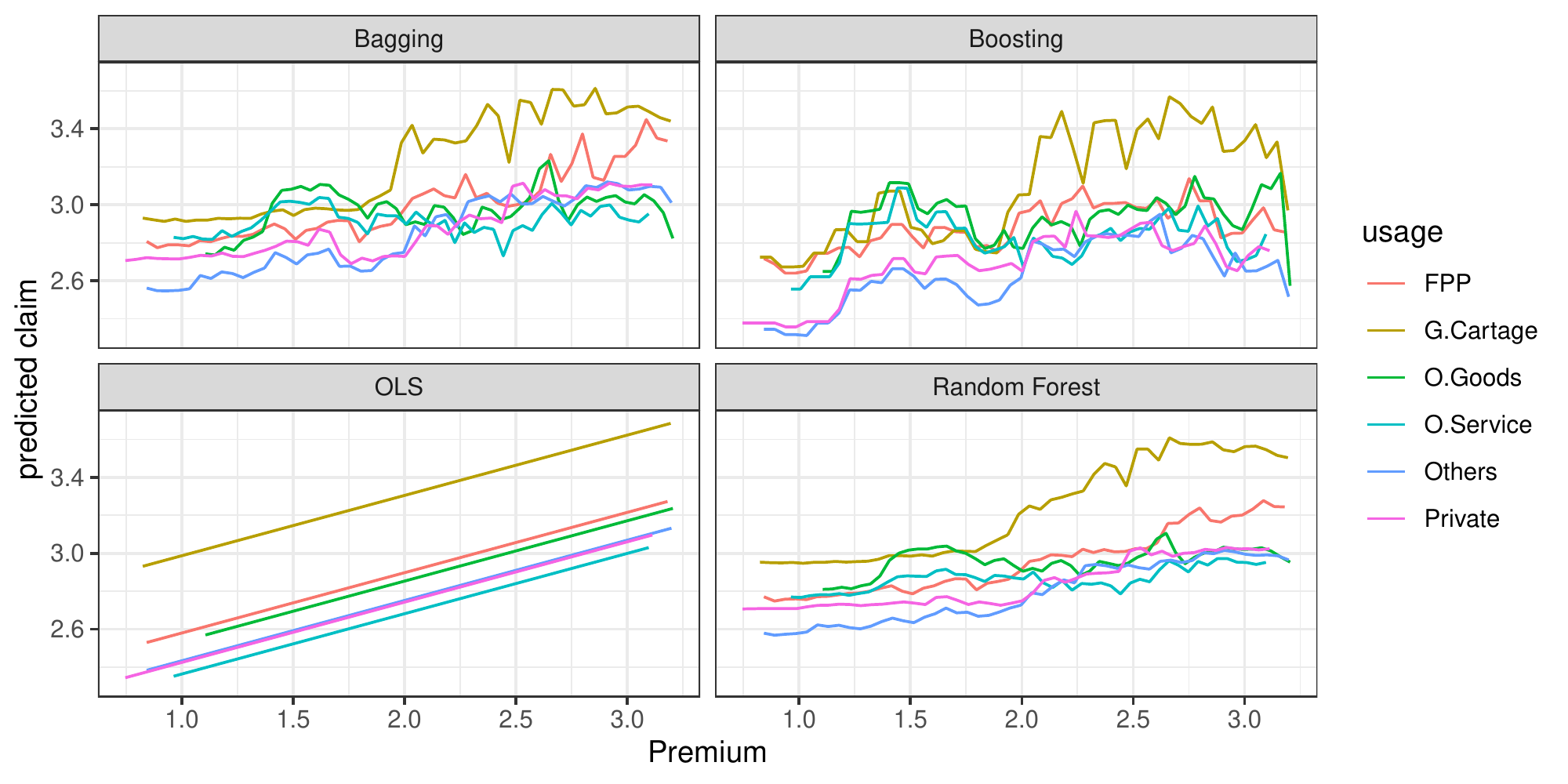}}
			\caption{Two-way partial dependence plots; the marginal effect of premium on the claims size for different groups of usage predictor after integrating out the other variables.}
			\label{fig:pdp}
		\end{center}
	\end{figure}
	
	Figure \ref{fig:pdp} shows a separate partial dependency function for each group of usage predictor. Because one-way partial dependency plots display one predictor at a time, they are valid only if the predictor of interest does not interact strongly with other predictors. However, interactions are common in actual practice; in these cases, we can use higher-order (such as two- and three-way) partial dependence plots to check for interactions among predictors. For example, Figure \ref{fig:pdp} shows an interaction between premium and usage predictors.
	
	The two-way plot shows that vehicles used for a general cartage with a high premium (more than 2 in ensemble methods) have much higher expected claims size compared to the vehicles with other usages. This interaction would have not apparent in the one-way plot.
	
	\subsection{Methods Comparison}
	In this application, the predictor variable is represented by a collection of quantitative and qualitative attributes of the vehicle and the response is the actual claims size. Given a collection of $N$ observations $\left\{(x_i,y_i); i = 1, \dots, N \right\}$ of known $(x,y)$ values, the goal is to use this data to obtain an estimate of the function that maps the predictor vector $x$ to the values of the response variable $y$. This function can then be used to make predictions on observations where only the $x$ values are observed. Formally, we wish to learn a prediction function $x \mapsto \hat{f}(x)$ that minimizes the expectation of some loss function $\mathcal{L}(\hat{f}(x), y)$ over the joint distribution of all $(x, y)$-values, that is
	
	\begin{equation}\label{eq:loss}
		\hat{f}(x) = \argmin_{f(x)} E \left[\mathcal{L}(f(x), Y) | X = x\right].
	\end{equation}
	
	In finite samples, we evaluate the performance of $\hat{f}$ with the Mean Square Error (MSE), that is
	
	\begin{equation}\label{eq:mse}
		MSE = \frac{1}{n'} \sum_{i=1}^{n'}\left(\hat{f}(x_i) - y_i\right)^2
	\end{equation}
	where $\hat{f}(x)$ is a fitted regression function on the test data set $\left\{x_i\right\}_{i=1}^{n'}$ and $y$ is the observed response variable.
	
	\begin{figure}[H]
		\begin{center}
			\includegraphics[width=1\textwidth]{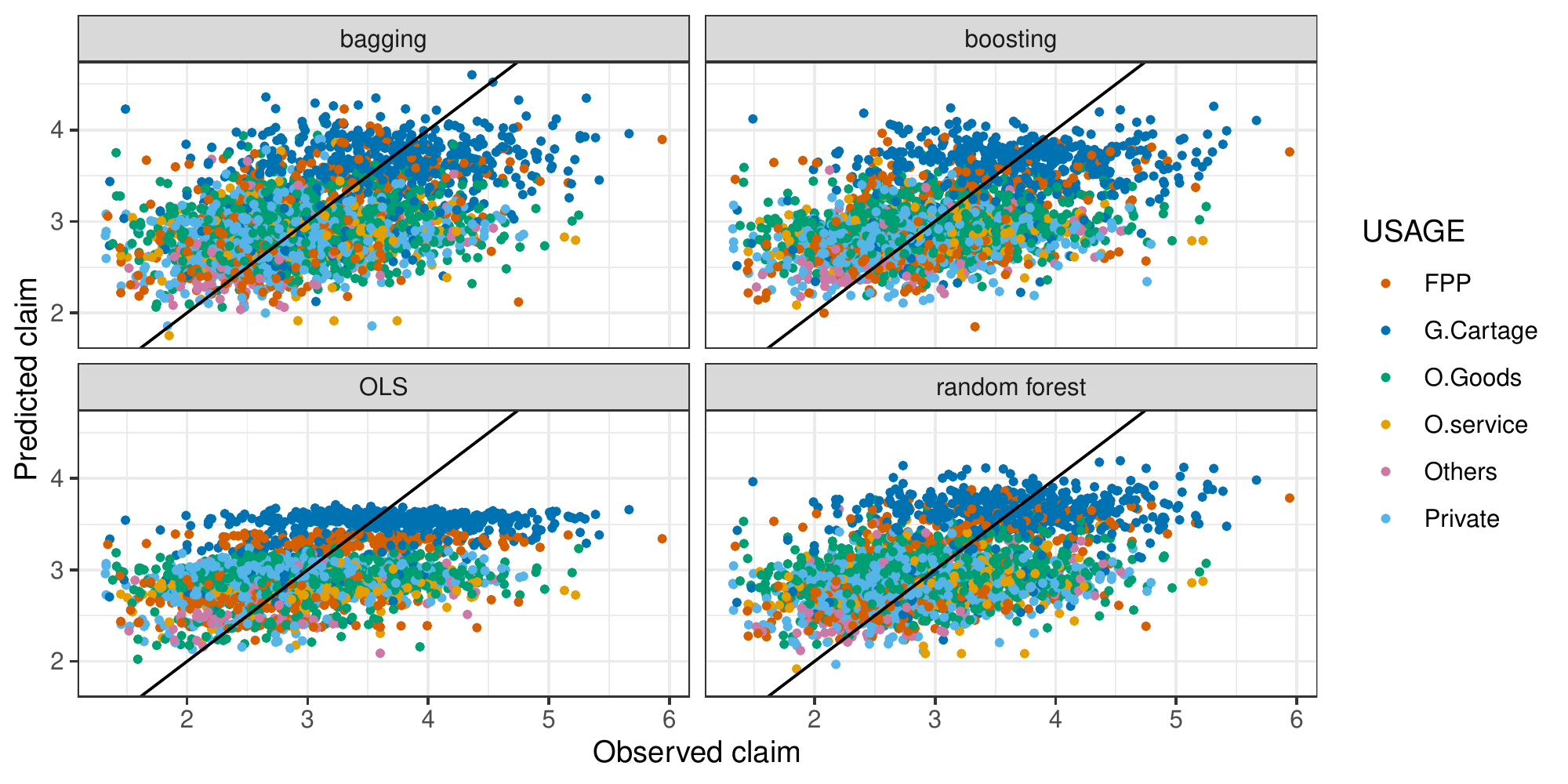}
			\caption{Observed against predicted claims size on log scales}
			\label{fig:obs_pred}
		\end{center}
	\end{figure}
	
	For statistical modeling purposes, we first partitioned the data into train $(70\%)$ and test $(30\%)$ data sets. The train set was used for exploratory data analysis, model training and selection, and the test set to assess the predictive accuracy of the selected method. The training data goes from years 2011 to  2015 and from 2017 to 2018, while the test data is observations from the year 2016.
	
	Regarding performance of the methods, it can be clearly seen that OLS method is predicting all the claims less than USD $10^4$ even though some observed claims are even larger than USD $10^5$. However in case of ensemble methods, they could predict claims size beyond $10^4$.  Both OLS and ensemble methods underestimated high claims size, but the underestimation is higher in OLS.
	
	\section{Conclusion} \label{conclusion}
	
	Ensemble methods are well established algorithms for obtaining highly accurate classifiers by combining less accurate ones. This paper has provided a brief overview of methods for constructing ensembles and reviewed the three most popular ones, namely bagging, random forest and gradient boosting. The paper has also provided some results from application of ensembles on real vehicle insurance dataset to address some problems of insurance companies. In the application section, the predictors are ranked according to their importance in predicting claims size, and the relationships between claims size and some predictors are assessed. The ­performances of non-ensemble (OLS) and ensemble learning algorithms (bagging, random forests and gradient boosting) are evaluated in terms of RMSE. Accordingly, the ensemble learning techniques outperformed the OLS. Thus, this study suggests that ensemble learning techniques can outperform non-ensemble techniques. Moreover, the three ensemble algorithms performed similarly.

	\bibliography{myreferences}
	\bibliographystyle{abbrvnat}
	
\end{document}